\pdfoutput=1

\documentclass[11pt]{article}

\usepackage[]{acl}

\usepackage{times}
\usepackage{latexsym}

\usepackage[T1]{fontenc}
\usepackage[utf8]{inputenc}

\usepackage{microtype}

\usepackage{inconsolata}

\usepackage{todonotes}
\usepackage{booktabs}
\usepackage{amssymb}
\usepackage{multirow}
\usepackage[font=small,belowskip=-5pt,aboveskip=3pt]{caption}
\usepackage{color}
\usepackage{soul}

\title{Parameter Efficient Tuning Allows Scalable Personalization of LLMs for Text Entry: A Case Study on Abbreviation Expansion}

\author{Katrin Tomanek, Shanqing Cai, Subhashini Venugopalan \\
  Google LLC}

\begin{document}
\maketitle
\begin{abstract}

Abbreviation expansion is a strategy used to speed up communication by limiting the amount of typing and using a language model to suggest expansions. 
Here we look at personalizing a Large Language Model's (LLM) suggestions based on prior conversations to enhance the relevance of predictions, particularly when the user data is small ($\approx$1000 samples). 
Specifically, we compare fine-tuning, prompt-tuning, and retrieval augmented generation of expanded text suggestions for abbreviated inputs. %
Our case study with a deployed 8B parameter LLM on a real user living with ALS, and experiments on movie character personalization indicates that
(1) customization may be necessary in some scenarios and prompt-tuning generalizes well to those, (2) fine-tuning on in-domain data (with as few as 600 samples) still shows some gains, however (3) retrieval augmented few-shot selection also outperforms fine-tuning. (4) Parameter efficient tuning allows for efficient and scalable personalization.
For prompt-tuning, we also find that initializing the learned ``soft-prompts'' to user relevant concept tokens leads to higher accuracy than random initialization.

\end{abstract}

\section{Introduction}
Language models have long been used to reduce keystrokes and aid text entry in smart keyboards. This work looks at models for such keyboard applications in Augmentative and Alternative Communication (AAC) devices, in particular those for users with severe motor impairments, e.g., people living with amyotrophic lateral sclerosis (ALS) who communicate through eye-gaze typing. Recent advances in the generative capabilities of large language models (LLMs) can help significantly accelerate communication for such users. Prior studies~\cite{adhikary2021accelerating,cai2022context,shen2022kwickchat} proposed techniques for abbreviation expansion, where a user types short keywords or abbreviated phrases consisting of the initial letter of each word and an LLM is used to generate the fully-expanded sentence. %
Including the conversation context~\cite{wisenburn2008aac,gorman2021structured} was shown to further improve the accuracy of the predictions. In this work we explore personalization, another dimension that can improve the relevance of the predictions to fit a user's vocabulary and language style. %

In many real-world applications personalization plays an important role in enhancing the relevance of the suggested options and the quality of the user experience \cite{valencia2023less}.
However, very little data is available to adapt a model to a given user, and larger models increase the risk of overfitting. %
Additionally, it remains unclear how to scale the approach to multiple users given the high cost of LLM checkpoint storage and serving.

With these challenges in mind, our work evaluates three approaches to personalizing LLMs for abbreviation expansion as used by eye-gaze typers. Specifically we consider a pre-trained decoder-only LLM tuned for dialog~\cite{roller2020recipes,thoppilan2022lamda}.
We further fine-tune the model on the abbreviation expansion task on data derived from dialog datasets. We then compare personalizing this fine-tuned LLM on user-specific data via (1) fine-tuning the entire model, (2) augmenting the LLM's context by retrieving similar conversations from the user's history, and (3) parameter efficient prompt-tuning~\cite{lester2021power}. Overall, prompt-tuning performed best and retrieval augmented in-context learning (RA-ICL) also outperformed fine-tuning.

\section{Related Work}

\subsection{Language models for text-entry.}
Using language models to expand abbreviated inputs for text-entry has been well studied and different schemes of abbreviation have been proposed such as, using just context words~\cite{demasco1992generating}, discarding vowels~\cite{shieber2007abbreviated}, and additionally omitting repeated consonants~\cite{willis2005probabilistic}, flexible letter saving schemes~\cite{adhikary2021accelerating,gorman2021structured}, and expanding from a bag of words~\cite{shen2022kwickchat}.  
Our study focuses on abbreviation expansion used by eye-gaze typers living with severe motor impairments. Given our goal to significantly reduce the number of keystrokes, %
we consider a form of word-initial abbreviation similar to ~\citet{cai2022context} where just the initial characters of the words %
are typed and an LLM predicts the full sentence. The current study focuses on personalizing such a model to a user, which has been less studied. %

\subsection{LLM prompt engineering.}
LLMs have shown remarkable capabilities in understanding and performing tasks with few-shot~\cite{brown2020language} examples. 
However, the tokenization used in LLMs makes our task of generating expansions from single characters somewhat hard for the models. 
Due to this reason and to enable personalization, 
we focus on  Parameter Efficient Fine-Tuning (PEFT)~\cite{lester2021power}, and retrieval augmented generation (RAG)~\cite{mialon2023augmented}. PEFT learns a small set of additional parameters while keeping the weights of the original LLM frozen. 
Many PEFT methods have been proposed in recent years. 
In case of adapters~\cite{houlsby2019parameter} and Low-Rank Optimization (LoRA)~\cite{hu2021lora} these parameters are interspersed at different transformer layers of the model. Other methods such as, Prompt-tuning~\cite{lester2021power}, Prefix-tuning~\cite{li2021prefix}, and P-tuning~\cite{liu2021gpt} restrict the parameters to the input prompt tokens. 
We use prompt-tuning~\cite{lester2021power} which append parameters to the token embeddings. We also compare this to retrieval augmentation for ICL~\cite{rubin2021learning} where %
a dense retriever is used to select
relevant data point(s) that are then added as context to a generative answering model. While most RAG studies~\cite{mialon2023augmented} train the retriever or the generator, we keep both of these pre-trained models frozen. %
Specifically, we use the retrieved context to create more relevant few-shot examples specific to the input query. %

\section{Tuning and personalization}
\vspace{-0.2cm}
Our broad approach consists of taking a pre-trained LLM, performing supervised fine-tuning for the abbreviation expansion task, and then personalizing the model on user data by means of further fine-tuning, prompt-tuning, or retrieval augmented in-context few-shot generation. For the pre-trained model, we start with an 8B parameter decoder-only LLM. %
This model is pre-trained on the C4 dataset~\cite{raffel2019exploring} and tuned for dialogs~\cite{roller2020recipes,thoppilan2022lamda}. We then fine-tune it further for abbreviation expansion on sentences from conversations and associated word-initial abbreviated text. We follow prior works~\cite{cai2022context} and experiment with different learning rates, and use a constant rate during fine-tuning %
and select the best based on a validation set. We refer to this as the base-AE model. %
We explore 3 strategies for personalization.

\subsection{Fine-tuning on user data.}
We follow the same fine-tuning recipe on user data as with the base-AE model. %
The tuning itself is fast since the amount of user data is small, and we avoided overfitting by monitoring performance on the validation set. %
We experimented with learning rates 1e-5, 1e-6, and 5e-5 and found 5e-5 to work best (see App. Tab.~\ref{tab:lr_ft}).

\subsection{Few-shot and Retrieval Augmented In-Context Learning (RA-ICL)}
Another way to personalize an LLM is to provide it with few-shot examples to allow for in-context learning (ICL). Performance with ICL can vary significantly with few-shot examples~\cite{zhao2021calibrate}. Hence, in addition to typical few-shot examples, we also investigate a retrieval-augmented few-shot setup. This is similar to works that retrieve from databases to augment LLMs~\cite{mialon2023augmented} but we use existing pre-trained models for retrieving and generating, and keep them frozen. %
For the retriever, we use a pre-trained 11B Sentence-T5~\cite{ni2021sentence} and generate embeddings of the abbreviated inputs from the user conversations. Given a new input, we embed it and use Euclidean distance to retrieve the nearest neighbor queries and the corresponding expansions. We use this retrieved context to create relevant, query-specific few-shot examples with which we prompt the LLM. %

\subsection{Prompt-tuning}
We also investigate prompt-tuning~\cite{lester2021power} for personalization. The basic idea is to extend few-shot prompting and use substantially more in-context examples to learn ``soft-prompts'' in the input embedding layer specifically suited for the task at hand. %
We choose the length of the soft prompt and initialize the tokens. 
For tuning, we correspondingly add new learnable parameters to the model's embedding matrix that are updated using back propagation, keeping the original LLM weights frozen.
The number of learned parameters is a product of the length of the soft-prompt and dimensions of the embedding weights. The learned soft-prompts are saved and passed along with each user query to the LLM during inference. This approach allows a single LLM to be served, and the soft-prompt to be swapped for different users (see Sec.~\ref{sec:disc}). The soft-prompts themselves can be tuned on varying amounts of data, and are effective in low data settings~\cite{lester2021power}. 
We train with a warm-up
learning rate schedule with 1000 warm up steps to a peak of 0.1 followed by linear decay. We use small batch sizes of 16 for training and limit training to 20k steps. We experiment with different prompt lengths and initialization strategies, and choose the best checkpoints based on validation set accuracy. 

\section{Dataset} 
\vspace{-0.2cm}
\subsection{Abbreviation Expansion Base Model}
\label{sec:ae-base-data}
To fine-tune the LLM for the abbreviation expansion task, we need pairs of abbreviated phrases and the full expanded text. We use the data from \citet{cai2022context} where they prepare paired sentences and abbreviated inputs from four dialog datasets: crowd-sourced Turk Dialogues~\cite{vertanen2017towards}, DailyDialog~\cite{li2017dailydialog}, %
the Cornell Movie Dialogues~\cite{danescu2011chameleons} from movie scripts, and Turk AAC dataset~\cite{vertanen2011imagination} of conversations collected with AAC users in mind.
The model fine-tuning is done with a constant low-learning rate (0.01) using the AdaFactor optimizer~\cite{shazeer2018adafactor} on over 237,000 examples derived from the dialog datasets.

\subsection{Personalization Dataset}
\vspace{-0.1cm}
A model trained on generic dialog datasets may not fit the communication needs of all in terms of preserving their style, and vocabulary including proper nouns. Our work is motivated to increase the autonomy and self-expression of AAC users with motor and speech impairments and deploy our abbreviation expansion model for their daily usage.
This is also a case where a generic models' training data is also lacking in terms of conversations around caregiving and health. 
Hence, our personalization dataset was collected from a person living with ALS with informed consent from the user and the conversation partners. They use eye-gaze text entry for everyday communication. They type on a virtual keyboard into the text editor of a text-to-speech (TTS) software and activate the audio to "speak" out the contents. Private and sensitive content was redacted prior to obtaining the data for research. The data was partitioned chronologically, and repetition was removed from the validation and test portions resulting in 630 (train), 194 (val.) and 224 (test) samples. %

\subsection{Movie character personalization}

Outside of the real deployment scenario, we also examined other conversation datasets where personalization can be studied without affecting user privacy.
Characters in movies and TV series tend to have certain quirks and personalities and make for a great test bed for evaluating personalization of spoken dialogues. Thus, to evaluate the need for customization and scalability of the approach, we performed additional experiments on conversations from the Cornell Movie Dialogs dataset~\cite{danescu2011chameleons} test set.  For our experiments, we selected 10 movies with very high ratings (with atleast 5k votes on ImDb). From each movie, we chose 1 character and all their conversations from the movie for personalization. Each character had over a hundred conversations in the movie (range 104 to 344, with a mean of 198.4 and median of 209 conversations). Similar to our AAC personalization dataset we did a time-based split of the data to get train, val., and test splits. 

\section{Experiments and Results}
\textbf{Experimental setup.} For all experiments, we sample 128 responses from the model with temperature 1.0, sort based on frequency of predictions and select the top-5. We report results on the average (and $\pm$ std. dev.) of 3 runs unless specified otherwise. %
The metrics we use are \textbf{Accuracy} to measure exact match of the full sentence expansion, and \textbf{BLEU} score~\cite{papineni2002bleu} to consider partial credit, both measured on the top-5 predictions (noted as \textbf{@5}).
\subsection{Prompt-tuning is best for Personalization}
Table~\ref{tab:eval_main} compares the performance of the different personalization approaches on the real user data. We note that the base-AE model achieved a top-5 accuracy of 68.3\% on the abbreviation expansion test set, however from Tab.~\ref{tab:eval_main} we can see that it only gets an accuracy of 22.5\% on the user personalization test set %
highlighting the difference between the user data distribution and the training distribution, and making a strong case for personalization for AAC users. Fine-tuning on user data helps, and retrieval for ICL does even better, however %
 prompt-tuning results in the best performance.

\begin{table}[h]
\small
\begin{center}
\resizebox{\columnwidth}{!}{
\begin{tabular}{ c | c | c | c}
  \hline
   Model & personalized & Accuracy@5 & BLEU@5 \\ 
   \hline \hline
  base-AE & $\times$ & 22.5 & 31.8 \\
  ICL & $\checkmark$ &  22.8 & 34.9 \\
  Fine-tuned & $\checkmark$ & 26.5 & 34.3 \\
  RA-ICL & $\checkmark$ &  30.3 & 39.1 \\
  Prompt-tuned & $\checkmark$ & \textbf{38.8} & \textbf{47.5} \\
  \hline
\end{tabular}
}
\end{center}
\caption{Accuracy (exact-match of full sentence) and BLEU score of top 5 predictions of the different approaches on the personalization test set.
\label{tab:eval_main}}
\vspace{-0.2cm}
\end{table}

\subsection{Soft prompt initialization matters}
We experimented with different soft-prompt lengths, learning rates, and soft-prompt initialization strategies. We tried soft-prompt lengths of 10, 25, and 100 tokens all initialized randomly. Recall that increasing the prompt lengths increases the number of learned parameters. In our case, we found higher prompt lengths led to more training instabilities. We found a length of 10 to work best. Fixing the prompt length as 10, we experimented with learning rates of 0.1, 0.2, and 0.3 and found 0.1 to work best (in App. Tab.~\ref{tab:lr_pt}).
\begin{table}[h]
\small
\begin{center}
\resizebox{\columnwidth}{!}{
\begin{tabular}{| l | c | c | c}
  \hline
  Soft-prompt Initialization & Accuracy@5 & BLEU@5 \\ 
   \hline \hline
  Random                & 32.7 $\pm 3.2$ & 43.6 $\pm 2.3$ \\
  LLM vocab. sampled    & 33.9 $\pm 0.4$& 43.2 $\pm 1.8$ \\
  User vocab. sampled   & 32.6 $\pm 1.6$ & 41.0 $\pm 1.9$\\
  User relevant concepts& \textbf{36.8} $\pm 1.9$ & \textbf{45.9 } $\pm 1.4$ \\
  User concept antonyms & \textbf{36.4} $\pm 0.3$& \textbf{46.2} $\pm 4.3$ \\
  \hline
\end{tabular}
}
\end{center}
\caption{Initializing soft-prompts with proper nouns and concepts from the user's data performs best.
\label{tab:prompt_init}}
\end{table}

The thing that made the biggest difference though was the choice of initialization for the soft-prompt token embeddings, which can be seen in Table~\ref{tab:prompt_init}. We examined 5 strategies, (1) random initialization, (2) sampling from the top 5k words in the LLM's vocabulary, (3) sampling from the top 25 most common English words in the user's vocabulary, (4) hand-selecting proper names and concepts relevant to the user (e.g. ALS) and (5) selecting words that are related but might be considered antonyms of the user concepts (e.g. Parkinsons). We found the initialization that relied on the user concepts to perform significantly better. Analogous to what is suggested in \citet{lester2021power}, perhaps these tokens are the ones the base model is most uncertain about, and hence boosts their chance of appearing in the predictions when prompt-tuned.

\subsection{Fine-tuning hampers generalization}
\begin{figure}[h]
    \centering
    \includegraphics[width=0.8\columnwidth]{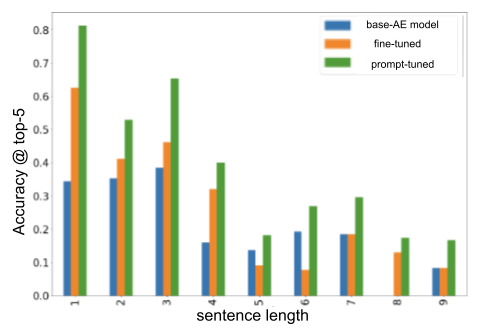}
    \caption{Performance of the different approaches on conversation sentences of different lengths. On longer sentences, the prompt-tuned (green) and base non-personalized model (in blue) can outperform fine-tuning highlighting their capacities to generalize to the long tail of complex sentences.}
    \label{fig:pers_2}
\vspace{-0.2cm}
\end{figure}

Fig.~\ref{fig:pers_2} slices performance of the models based on the length of the sentences. The performance of all models degrade with increasing sentence length. %
However, the fine-tuned model generalizes poorly compared to the base-AE model in some cases (noticeable at lengths 5 and 6). This also highlights the difficulty with fine-tuning large models on very small datasets.

\begin{table}[!h]
\small
\begin{center}
\resizebox{\columnwidth}{!}{
\begin{tabular}{| l | c | c | c}
  \hline
  ICL (4-shot) strategy & Accuracy@5 & BLEU@5 \\ 
   \hline \hline
  Random 4-shot         & 22.0 $\pm 0.9$ & 30.8 $\pm 1.4$ \\
  Hand-crafted 4-shot   & 21.9 $\pm 0.5$ & 33.3 $\pm    0.7$ \\
  Retrieval augmented (RA-ICL) & 30.2 $\pm 0.3$ & 38.5 $\pm 0.6$ \\
  \hline
\end{tabular}
}
\end{center}
\caption{Comparing different few-shot selection strategies. Retrieval augmented ICL works best.
\label{tab:icl_strategy}}
\end{table}

\subsection{Retrieval augmented few-shots help}
Table~\ref{tab:icl_strategy} presents results for in-context learning where 4-shot examples are selected using different strategies: (1) random selection from the training set, (2) hand-crafted examples containing proper names of people that the user communicates with, and (3) retrieval-augmented few shots, where 4 nearest neighbor examples (in the embedding space of abbreviations) are selected based on each test query. RA-ICL outperforms other strategies by a large margin.

\begin{table*}[t]
\centering
\resizebox{\linewidth}{!}{
\begin{tabular}{cccccccc}
\toprule
\textbf{Movie-id} & \textbf{character-id} & \textbf{Character-Name} & \multicolumn{2}{c}{\textbf{Non-personalized base-AE}} & \multicolumn{2}{c}{\textbf{Personalized (prompt-tuned)}} & \textbf{Personalization} \\
\cmidrule(r){4-5} \cmidrule(l){6-7}
& & & Acc.~@5 & BLEU @5 & Acc.~@5 & BLEU @5 & \textbf{rel. benefit (\%)} \\
\midrule
m106 & u1612 & JACOB & 62.75 & 67.03 & 56.86 & 65.13 & - \\
m119 & u1799 & GEORGE & 50.00 & 59.18 & 56.25 & 62.46 & 13\% \\
m126 & u1916 & ANDERTON & 44.12 & 55.66 & 38.24 & 55.74 & - \\
m140 & u2157 & BABE & 60.00 & 69.52 & 46.67 & 62.93 & - \\
m148 & u2299 & NANCY & 41.67 & 52.67 & 41.67 & 51.40 & - \\
m203 & u3105 & MICHAEL & 61.90 & 59.60 & 47.62 & 45.32 & - \\
m274 & u4099 & ABBY & 77.78 & 77.78 & 77.78 & 77.78 & - \\
m324 & u4866 & SONNY & 62.86 & 71.53 & 65.71 & 72.97 & 5\% \\
m352 & u5310 & JACK & 50.00 & 59.18 & 56.25 & 62.46 & 13\% \\
m565 & u8329 & JEANNE & 61.54 & 70.61 & 64.10 & 71.24 & 4\% \\
\bottomrule
\end{tabular}
}
\caption{Performance comparison between Non-personalized base-AE and Personalized (prompt-tuned) models on movie character personalization. LLMs raise the bar on average performance indicating that customization may not always be necessary on certain conversation categories, though some users benefit from it. This is a contrast to the real AAC deployment scenario.}
\label{tab:movie-peft}
\end{table*}

\subsection{Customization is not always necessary}
We also evaluated the prompt-tuning approach on the movie character personalization dataset and report results in Table~\ref{tab:movie-peft}. We observe that:
(1) the base non-personalized model accuracies do seem to transfer reasonably well indicating that customization may not be necessary for conversation types similar to the training data distribution.
(2) 4 of the 10 speakers still benefit from personalization.
(3) the proposed \textbf{prompt-tuning approach offers a way to serve the same end-point, while optionally choosing to personalize results to some users}.
\begin{table*}[!t]
\begin{center}
\resizebox{\linewidth}{!}{
\begin{tabular}{ l | l | l | l | l }
  \hline
  \textbf{Error Type}  & \textbf{Abbreviation} & \textbf{Gold Expansion} & \textbf{Fine-tuned} & \textbf{Prompt-tuned} \\ 
   \hline \hline
  Unmatched Acronym                & s i l t r & \textcolor{blue}{\ul{sweet}} i love that robin & i love that robin & sweet i love that robin \\ \hline
  Overfitting to names    & g q d , r a m & great question dude , robin and mommy & \textcolor{red}{greg} q day, robin and \textcolor{red}{greg} & good q doc, robin and mommy \\ \hline
  Misses user style (often &  \multirow{2}{0.24\columnwidth}{w a d , o d y} & \multirow{2}{0.6\columnwidth}{what a \textcolor{blue}{\ul{dunce}} , okie dokie \textcolor{blue}{\ul{yall}}} & what about daddy , okie dokie & what a day , okie dokie \\
  \cline{4-5}
   when lacking context) & &  & wipe and dry , ok thanks & we are done , ok day yall \\
  \hline
\end{tabular}
}
\end{center}
\caption{Examples of some observed categories of errors. Words that the model \textcolor{blue}{\ul{misses}} are highlighted in \textcolor{blue}{\ul{blue}} in the Gold expansion, and errors in names are marked in \textcolor{red}{red}. (Proper names have been changed to preserve anonymity)
\label{tab:error_analysis}}
\vspace{-0.2cm}
\end{table*}

\subsection{Error Analysis}
In Table~\ref{tab:error_analysis} we share some examples of the categories of errors we observe comparing the fine-tuned and prompt-tuned results. Our analysis of the predictions show that the fine-tuned model tends to overfit to proper nouns in the user's training data, and often misses generating expansions for some of the characters in the acronym. On sessions where there is not enough user context, it can miss the user's style (e.g. the word contraction ``yall'' in row 4 of Table~\ref{tab:error_analysis} is less common in general text, but could be stylistic of a user).

\section{Discussion}
\label{sec:disc}
\vspace{-0.2cm}
\subsection{LLM Blind-spots.}
Abbreviation expansion may seem to be an easy task for current LLMs. However, our work focuses on abbreviations motivated to help users with severe disabilities, and hence pushes the limit of keystroke savings. %
Wherein, the task depends on recognizing individual characters/alphabets. Interestingly, it falls into what could be a "blind-spot" for the LLMs because the input tokenization schemes - meant to overcome a discrete vocabulary - may fall short in recognizing individual characters. This is now addressed practically e.g. for generating text in JSON format~\cite{GuaranteeValidJSONOutput}, using constrained decoding and following Backus-Naur Form (BNF) grammar.

\subsection{Data efficiency and scaling.}
Another point of discussion is how personalization can be performed on a small amount of data. Our experiments show that prompt-tuning leads to higher test accuracy than fine-tuning in limited data settings. %
Fine-tuning the full LLM for personalization not only generalizes poorly, but is also very expensive in terms of storing the personalized model weights. Prompt-tuning on the other hand only involves storing a very small set of weights (on the order of thousands) which would make it not only possible but also convenient to store these on users' personal devices. This also makes the approach more scalable since only a single model can be served, while clients can query it using different personalized soft prompts. Further, querying a prompt-tuned model incurs little additional inference latency, as the learned prompts and the user input are provided simultaneously to the model during inference.

\vspace{-0.2cm}
\section{Conclusion}
\vspace{-0.2cm}
Our work presents a case study on personalizing LLMs for the task of abbreviation expansion in the context of aiding eye-gaze typers with severe motor and speech disabilities to communicate faster. We fine-tuned an LLM on generic dialog data for the task and compared approaches to personalization using limited user data. We examined fine-tuning, parameter-efficient prompt-tuning, and retrieval augmented in-context learning, and find prompt-tuning to be the most elegant method for personalization in terms of its performance as well as its training data efficiency, small storage requirements, and ability to scale. Further, initializing the soft-prompts with concepts and terms relevant to the user resulted in better prompt-tuned personalized models. %

\section*{Limitations}
The effectiveness of personalization on real usage is difficult to study, since it deals with private and sensitive content. This difficulty is more pronounced when working with people with disabilities. This limits our work to a case study on real user data for personalization. Identifying interesting techniques to collect realistic personalization datasets, perhaps synthetic, can benefit the community significantly.

We also limit the extent of hyperparameter tuning, due to significant computation resource consumption of experiments. Though we are able to take advantage of settings shared in literature and open source code. Also, while our abbreviation expansion study and models are limited to English, it will likely translate well to languages with similar morphology, but that remains to be studied. Our references to related work in this space may be limited and further suggestions are welcome.

\section*{Ethics and Societal Impact}
Techniques that improve Augmentative and Alternative Communication (AAC) applications can significantly enhance quality of life, increase independence and social participation~\cite{calgari2013} of people living with communication and motor disabilities. %

A risk of abbreviation expansion is that, when the expansions are not exactly the ones that the user desires, they may be tempted to choose a near similar prediction leading to conveying content that may be less accurate, misinterpreted, or reflecting biases and stereotypes of the underlying models. While the goal of personalization is to mitigate these, some of the risks still remain. Hence there is still a subtle risk of reducing speaker's autonomy and authentic self-expression which people e.g. with ALS~\cite{kane2017cscw} value highly. Another risk is that of frequent incorrect predictions if personalization is poor for some users. This could increase effort required to edit minor errors, and inadvertently increase fatigue.

\bibliography{references}
\bibliographystyle{acl_natbib}

\newpage

\appendix

\section{Parameter selection}

\subsection{Fine-tuning learning rates}

\begin{table}[!h]
\small
\begin{center}
\resizebox{\columnwidth}{!}{
\begin{tabular}{| l | c | c | c}
  \hline
  Fine-tuning learning rate & Accuracy@5 & BLEU@5 \\ 
   \hline \hline
  5e-5         & 26.8 & 34.3 \\
  1e-6   & 25.4 & 34.7 \\
  1e-5 & 23.7 & 31.6 \\
  \hline
\end{tabular}
}
\end{center}
\caption{Comparing different learning rates for fine-tuning base-AE model on personalization data. (val set).
\label{tab:lr_ft}}
\end{table}

\subsection{Prompt-tuning learning rates}

\begin{table}[!h]
\small
\begin{center}
\resizebox{\columnwidth}{!}{
\begin{tabular}{| l | c | c | c}
  \hline
  Prompt-tuning learning rate & Accuracy@5 & BLEU@5 \\ 
   \hline \hline
  0.1         & 35.7 & 45.6 \\
  0.2   & 31.7 & 41.7 \\
  0.3 & 30.8 & 39.9 \\
  \hline
\end{tabular}
}
\end{center}
\caption{Comparing different learning rates for prompt-tuning base-AE model on personalization data. soft prompt length of 10 and random initialization (val set).
\label{tab:lr_pt}}
\end{table}

\section{Personalization Data}
\label{sec:appendix_user}
Our personalization dataset was collected with informed consent from a person living with ALS over a period of five months from late 2021 to early 2022.
We refer to the person with ALS as "the user". The user used a Tobii (R) eye-tracker and gaze-driven keyboard to enter text for daily communication. The gaze-typed text was output as speech audio through text-to-speech (TTS) software.
The user had full control over when to start and stop data collection. Private and sensitive content in the data was redacted by trained human curators prior before we ingested the dataset for research.

The relevant data used for this study consists of text transcripts of the user's TTS output. We split the data into three non-overlapping splits along the time axis in %
chronological order as train, validation and test, containing 630, 285, and 284 sentences, respectively.
We filter the validation and test split to preserve only the sentences with abbreviation length $\leq 10$, leading to 194 and 224 sentences, respectively. No filtering is done on the training split. As a result, the average abbreviation length in the train, validation, and test splits are $6.91\pm6.25$, $4.72\pm2.39$, and $5.05\pm2.74$, respectively ($\pm 1 SD$). The sentences belong to 122, 69, and 72 sessions, respectively, each session being a continuous period of conversation data collection.
The percentages of proper nouns among the words were $6.73\%$, $5.88\%$, and $8.61\%$ in the three splits, respectively.

\end{document}